\documentclass[10pt,twocolumn,letterpaper]{article}
\usepackage{cvpr}
\usepackage{times}
\usepackage{epsfig}
\usepackage{graphicx}
\usepackage{amssymb}
\usepackage{mathtools}
\usepackage{algorithm}  
\usepackage{algpseudocode}  
\usepackage{amsmath} 
\usepackage{setspace}
\usepackage{booktabs}
\usepackage{color,xcolor}
\usepackage{bbding}

\usepackage[pagebackref,breaklinks,colorlinks]{hyperref}
\usepackage[capitalize]{cleveref}
\crefname{section}{Sec.}{Secs.}
\Crefname{section}{Section}{Sections}
\Crefname{table}{Table}{Tables}
\crefname{table}{Tab.}{Tabs.}

\def\cvprPaperID{10194} 
\def\confName{CVPR}
\def\confYear{2022}

\begin{document}

\title{Full Transformer Framework for Robust  Point Cloud Registration with Deep Information Interaction }  

\author{Guangyan Chen$^1$, Meiling Wang$^1$, Yufeng Yue$^{1*}$, Qingxiang Zhang$^1$, Li Yuan$^2$
\\
$^1$ Beijing Institute of Technology
$^2$ National University of Singapore
}

\maketitle

\begin{abstract}
Recent Transformer-based methods have achieved advanced performance in point cloud registration by utilizing advantages of the Transformer in order-invariance and modeling dependency to aggregate information. However, they still suffer from indistinct feature extraction, sensitivity to noise, and outliers. The reasons are: (1) the adoption of CNNs fails to model global relations due to their local receptive fields, resulting in extracted features susceptible to noise; (2) the shallow-wide architecture of Transformers and lack of positional encoding lead to indistinct feature extraction due to inefficient information interaction; (3) the omission of geometrical compatibility leads to inaccurate classification between inliers and outliers. To address above limitations, a novel full Transformer network for point cloud registration is proposed, named the Deep Interaction Transformer (DIT), which incorporates: (1) a Point Cloud Structure Extractor (PSE) to model global relations and retrieve structural information with Transformer encoders; (2) a deep-narrow Point Feature Transformer (PFT) to facilitate deep information interaction across two point clouds with positional encoding, such that Transformers can establish comprehensive associations and directly learn relative position between points; (3) a Geometric Matching-based Correspondence Confidence Evaluation (GMCCE) method to measure spatial consistency and estimate inlier confidence by designing the triangulated descriptor. Extensive experiments on clean, noisy, partially overlapping point cloud registration demonstrate that our method outperforms state-of-the-art methods. Code is publicly available at \url{https://github.com/CGuangyan-BIT/DIT}.
\end{abstract}

\begin{figure}[htbp]
\setlength{\abovecaptionskip}{5pt}
    \centering
    \includegraphics[width=8.3cm]{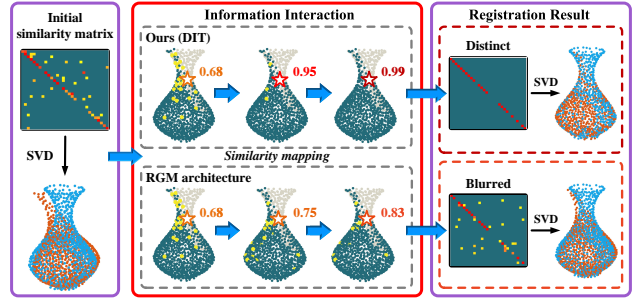}
    \caption{\textbf{The comparison between RGM \cite{fu2021robust} and our method (DIT) on partial-to-partial point cloud registration.} 
    Through deep information interaction, DIT obtains higher values in the similarity mapping for queried points (\textcolor{red}{\textbf{\FiveStarOpen}}). 
    Therefore, DIT generates a more distinct similarity matrix by improving the {discrimination} of features, and ultimately improves registration accuracy.}
    \label{fig_intro}
        \vspace{-13pt}
    \end{figure}

\section{Introduction}
  Point cloud registration   aims  to calculate a rigid transformation to align two point clouds, which is a key technology for  3D reconstruction, simultaneous localization and mapping (SLAM)  \cite{han2019real,deschaud2018imls}. In recent decades, point cloud registration has developed from model-based methods \cite{besl1992method,bouaziz2013sparse,jian2010robust,zhou2016fast,rusu2009fast}, and convolutional neural network (CNN)-based methods \cite{choy2020deep,gojcic2020learning,zhao2021centroidreg,aoki2019pointnetlk},  to recent Transformer-based methods \cite{wang2019deep,shi2021keypoint,fu2021robust,min2021geometry}. \textcolor{black}{These attempts have greatly increased the accuracy and robustness of point cloud registration by improving the defects of inefficient feature extraction and blurred mapping}. 
   
    The most widely known model-based method is iterative closest point (ICP) \cite{besl1992method},
    which iteratively alternates between establishing correspondences and calculating a transformation.
    However, ICP and its variants \cite{koide2021voxelized,tazir2018cicp,parkison2018semantic,chetverikov2005robust,serafin2015nicp,segal2009generalized} {tend} to converge to local minima due to the limited descriptive power of hand-crafted features. 
    Recently proposed CNN-based methods are able to learn rich and general features compared with model-based methods. 
    However, they
    only extract features from each point cloud separately \textcolor{black}{and are consequently unable to extract discriminative  features or identify common structures between two point clouds.} 
    Attention-based models such as Transformer \cite{vaswani2017attention},
   which were first applied in natural language processing (NLP)  \cite{devlin2018bert,brown2020language,krizhevsky2017imagenet}, have also shown  superior ability to extract features and aggregate information in computer vision tasks  \cite{huang2019ccnet,wang2020axial,
    carion2020end,ramachandran2019stand,dosovitskiy2020image,yuan2021tokens,yuan2021volo}.
    Inspired by the success of Transformers,
    recent works  \cite{shi2021keypoint,wang2019deep,wang2019prnet,fu2021robust,hertz2020pointgmm} investigated the advantages of Transformer models in point feature extraction.  Most of them utilized the attention mechanism to establish associations across two point clouds for information aggregation. \textcolor{black}{However, substantial gaps remain in terms of modeling global relations, enhancing feature richness, and detecting inliers:}
    \textcolor{black}{
    (1) current methods mainly add attention modules to a CNN framework for information aggregation, leading to the sensitivity to noise;
    (2) the insufficient associations established by shallow-wide Transformers and lack of positional information  prevent the model from {enhancing feature richness} and extracting distinct features;
    (3) the inlier detection {modules} did not consider spatial consistency of rigid transformations, resulting in low accuracy in removing outliers.}
   
  We argue that there is a key similarity between NLP and point cloud registration, namely, the need to establish associations between units representing the same content but with different expressions.  Motivated by this observation and the limitations of previous Transformer-based methods,   we propose a novel full Transformer framework named Deep Interaction Transformer (DIT), which takes advantage of the Transformer architecture to achieve  global receptive field and deep information interaction. The process of deep information interaction is shown in Fig. \ref{fig_intro}, which improves the discrimination of features. Experimentally, the proposed method is compared with extensive registration methods, indicating that the proposed method achieves superior performance.  The main contributions are four-fold:
\begin{itemize}
\setlength{\itemsep}{2pt}
\setlength{\parsep}{2pt}
\setlength{\parskip}{2pt}
  \item A Point Cloud Structure Extractor (PSE) is proposed to model global relations and integrate structural information. Concretely, Transformer encoders are adopted to model dependencies in the entire point cloud, enhancing the robustness to noise. In PSE, the Local Feature Integrator (LFI) is designed to structurize the point cloud, which addresses the limitation of Transformers in extracting structural features.
  
  \item A Point Feature Transformer (PFT) is proposed to {increase the richness of feature representation}. Specifically, deep-narrow Transformers are adopted to establish  comprehensive associations. In PFT, a positional encoding network is introduced to allow the model to  learn the relative position between points. 
 
\item A Geometric Matching-based Correspondence Confidence Evaluation (GMCCE) method is proposed to estimate the correspondence confidence based on geometric constraints. Specifically, a rotation-invariant triangulated descriptor is designed to measure geometrical compatibility.
\item The proposed DIT is a systematic framework that implements  above novel components {to improve the robustness to noise and discrimination of features.} The DIT
achieves superior performance in accuracy and robustness compared with state-of-the-art methods.

       \end{itemize}
   
      

      
\section{Related Work}
\subsection{Model-Based Registration Methods} 
 The most representative model-based method is the ICP algorithm \cite{besl1992method}, 
 which iteratively alternates between finding the closest points as correspondences and calculating {a} transformation based on the identified correspondences. 
 However, ICP and its variants \cite{segal2009generalized,rusinkiewicz2001efficient,pomerleau2015review,bouaziz2013sparse} often
 converge to local minima when the initial position is far from the global minimum. There is a large volume 
 of works \cite{eckart2018hgmr,campbell2016gogma,campbell2019alignment,maron2016point,rosen2020certifiably,izatt2020globally,yang2020teaser} that attempt to improve the robustness of ICP under poor initialization.
 In Gaussian mixture models (GMMs)\cite{jian2010robust},
  the registration problem is reformulated as the 
 alignment of two probability distributions.
 However, these methods still require  a warm initialization due to their nonconvex
 objective functions. 
 In globally optimal ICP (Go-ICP) \cite{yang2015go},
 the branch-and-bound (BnB)  method is applied
 to search over $SE(3)$ space to achieve global convergence, but the computational complexity is much higher than that of ICP. Fast global registration (FGR) \cite{zhou2016fast} relies on optimizing a global objective function
 to align two point clouds without any updating of correspondences. In addition, hand-crafted local features \cite{rusu2008aligning, salti2014shot, rusu2009fast} such as  fast point feature histograms (FPFH) 
 are also designed to  establish correspondences through feature matching. However, the accuracy and robustness of all of these  model-based methods are sensitive to partially visible point clouds and large initial errors.
 \begin{figure*}[ht]
 \setlength{\abovecaptionskip}{-0.13cm}
    \begin{center}
    \includegraphics[width=17.6cm]{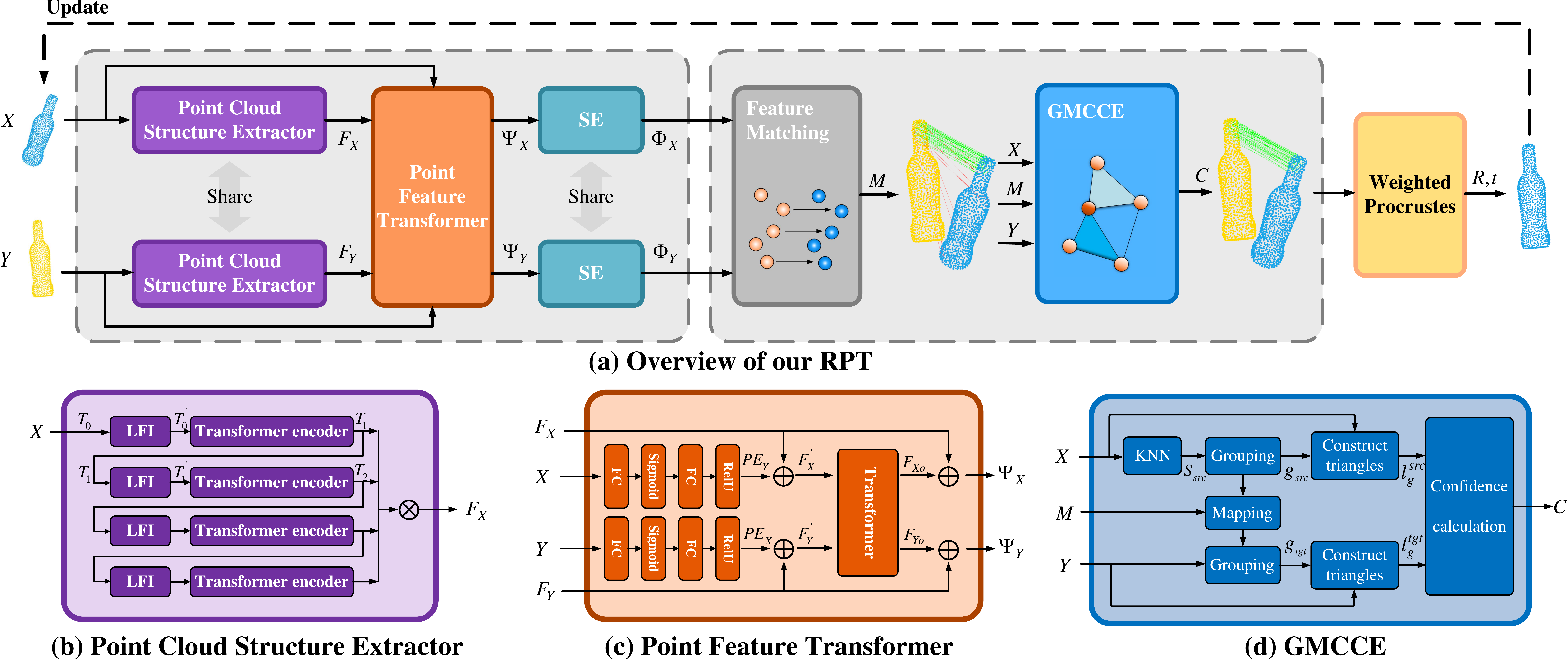}
    \end{center}
       \caption{\textbf{ (a) Network architecture of the Deep Interaction Transformer (DIT).}  DIT consists of three main components: (b)  Point Cloud Structure Extractor (PSE) and  (c) Point Feature Transformer (PFT) to extract features, where $\otimes$ denotes concatenation, $\oplus$ denotes matrix addition; (d) Geometric Matching-based Correspondence Confidence Evaluation (GMCCE) module to evaluate correspondence confidence. }
    \label{Overview}
\vspace{-12pt}
\end{figure*}

\subsection{{CNN-Based Registration Methods}}
The success of deep learning in point cloud processing \cite{zaheer2017deep,qi2017pointnet,qi2017pointnet++,phan2018dgcnn,choy2019fully} enables its application in point cloud registration. 
One pioneering  work is PointNetLK (PNetLK) \cite{aoki2019pointnetlk}, which extracts global features using PointNet \cite{qi2017pointnet} and applies the inverse compositional Lucas-Kanade (IC-LK) algorithm \cite{lucas1981iterative} to align two point clouds.  
PointNetLK Revisited (PNetLK\_R) \cite{li2021pointnetlk} has been proposed to circumvent the numerical instabilities of PointNetLK using analytical Jacobians.
However, since PointNet cannot aggregate the information from two point clouds, these two methods are sensitive to partially visible point clouds.
Deep Gaussian mixture registration (DeepGMR) \cite{yuan2020deepgmr} relies on a neural network  to predict the GMM parameters
and recover the optimal transformation. However, due to the independence of the feature extraction from  two point clouds, the features extracted by DeepGMR are indistinct.
A robust point matching network (RPM-Net) \cite{yew2020rpm} combines the Sinkhorn method with deep learning to establish soft correspondences from hybrid features, thereby enhancing the robustness to noise.
In summary, these methods extract features from each point cloud separately
and lack information interaction between the source and target point clouds, which is inefficient in  discriminative feature extraction and   contextual information aggregation, especially in partial-to-partial point cloud registration tasks.

\subsection{{Transformer-Based Registration Methods}}
Inspired by the success of Transformers in NLP and computer vision, researchers have begun to apply Transformers to extract contextual information between two point clouds. 
Deep closest point (DCP) \cite{wang2019deep} extracts features using dynamic graph CNN (DGCNN) \cite{phan2018dgcnn} and utilizes a Transformer \cite{vaswani2017attention} to aggregate  information. However, DCP lacks an overall understanding of {the} point cloud due to its local receptive field, which leads to sensitivity to noise. A multiplex dynamic graph attention network (MDGAT) \cite{shi2021keypoint} 
dynamically constructs a multiplex graph based on an attention mechanism.
A geometry guided network \cite{min2021geometry} encodes global  and local  features based on a self-attention mechanism with a fully connected graph.
{The recent robust graph matching (RGM) method \cite{fu2021robust} adopts a Transformer to aggregate information by generating soft graph edges. {However, the edge adjacency matrices are {indistinct} due to the shallow-wide architecture of the Transformer, which leads to a decrease in registration accuracy.}}
In summary, these methods mainly focus on modeling local relations by directly adopting convolution encoders,  which prevents them from modeling global relations.
Furthermore, the information interaction in these methods is inefficient due to the shallow-wide Transformer architecture and the lack of positional encoding.

\section{The Proposed Deep Interaction Transformer} 
 The point cloud registration problem aims to find a transformation to align two  point clouds. Given two point clouds $X = \left\{{x_{1},x_{2},...,x_{N}}\right\}  \subseteq R^{3}$ and $Y = \left\{{y_{1},y_{2},...,y_{M}}\right\}  \subseteq R^{3}$, which are denoted by $src$ and $tgt$, respectively, the objective is to estimate  a rotation matrix $R \in  SO(3)$ and a translation vector  $t \in R^3$ to map $src$ to $tgt$.

The overall pipeline of DIT is shown in Fig. \ref{Overview}(a). During training, the registration pipeline begins by extracting pointwise features $F_X$ and $F_Y$ from $src$ and $tgt$ separately using PSE. Then, deep information interaction is conducted by PFT to learn contextual information and extract discriminative features $\Phi _{X}$ and $\Phi _{Y}$. These features are matched to establish putative correspondences $M \{x_i,y_j\}$.  Finally, the weighted Procrustes module estimates the optimal transformation $\{ R,t\} $ to align the two point clouds based on the established correspondences $M$ and the similarity $S$ between the corresponding feature vectors $\{ \Phi _{x_i},\Phi _{y_j} \}$. During testing, \textcolor{black}{the GMCCE module is introduced to evaluate the correspondence confidence $C(x_i,y_j)$,} and then the weighted Procrustes module estimates the optimal transformation based on the confidence $C(x_i,y_j)$ instead of {the feature vector similarity}.


\subsection{Point Cloud Structure Extractor}
{Since the previous Transformer-based methods  mainly employ features from CNN, which cannot model global relations.
Therefore, the PSE module is designed to enhance the robustness to noise by modeling dependencies in the entire point cloud.}
Fig.  \ref{Overview}(b) shows the PSE architecture,
which consists of two types of components: LFIs 
and Transformer encoders {\cite{ba2016layer}.}  

\begin{figure}[t] 
\setlength{\abovecaptionskip}{5pt}
    \centering
    \includegraphics[width=8.2cm]{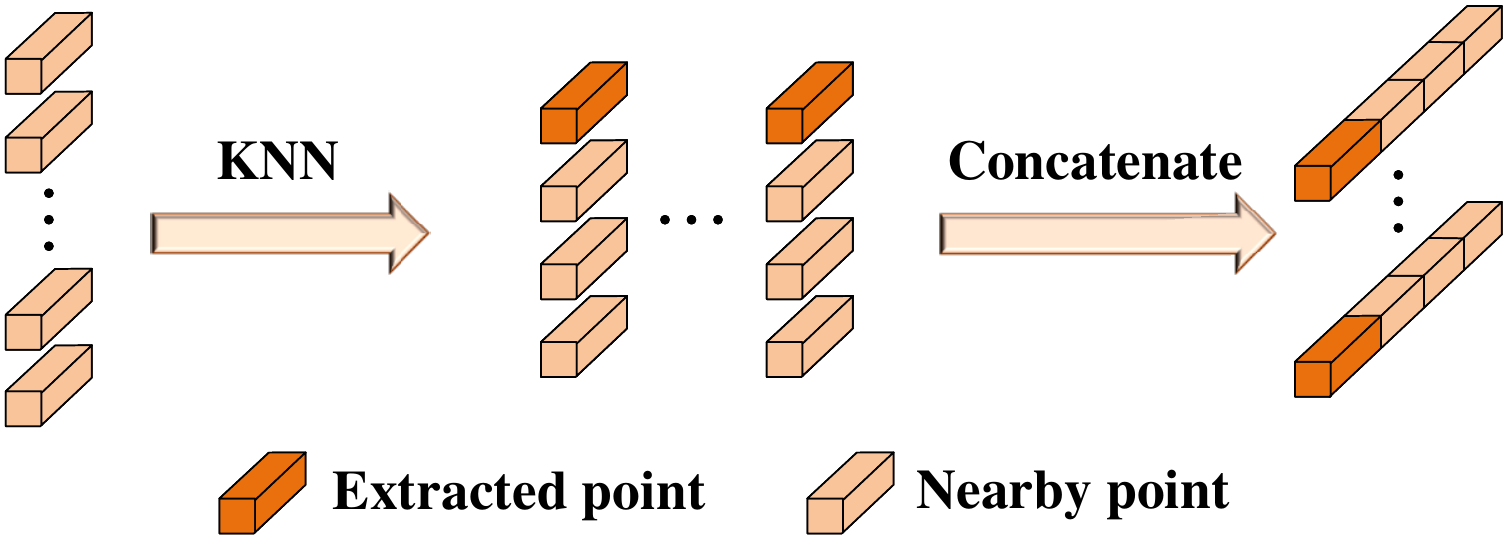}
    \caption{\textbf{Local Feature Integrator (LFI).} 
   KNN is applied to search for nearby points,  then the feature vectors of nearby points {and the extracted point are concatenated.}}
    \label{P2P_F}\vspace{-12pt}
    \end{figure}

{To overcome the limitation of Transformers in structural feature extraction \cite{dosovitskiy2020image,yuan2021tokens},
the LFIs are designed to progressively structurize the point cloud.} 
As detailed in Fig. \ref{P2P_F},
to identify the characteristics of the neighboring structures,
the {$n_{\rm th}$} LFI ($n\! = \! 1,\! \dots \! ,N_l$) searches for the local point cloud $P_n$ that contains the $k$ nearest points for each point in $X$, where $N_l$ denotes the number of LFI layers. Specifically, the LFI applies the k-nearest neighbors (${\rm KNN}$) method in geometric space instead of feature space to reduce the computational expense. 
Then, the {$n_{\rm th}$} LFI integrates structural information by concatenating ($\rm Concat$)
feature vectors $T_{Pn}^{i}$ of the points in $P_n$ to
construct integrated feature vectors $T_{n}^{'}$:
\begin{equation} \label{P2P} \small
\setlength{\abovedisplayskip}{5pt}
\setlength{\belowdisplayskip}{5pt}
    T_{n}^{'}  = {\rm Concat}([ T_{Pn}^{1},T_{Pn}^{2},\cdots ,T_{Pn}^{k}]).
\end{equation}

{With the structural features $T_n^{'}$, the Transformer encoder is adopted to model global relations in the point cloud \cite{zhao2021point}.}
Each Transformer encoder consists of a multilayer perceptron
(${\rm MLP}$), layer normalization (${\rm LN}$), {and the {multi-head} self-attention operation (${\rm MSA}$), which is based on {multi-head} attention (${\rm MA}$).
${\rm MA}$ is formulated as}
\begin{equation} \label{self-Attention} \small
\setlength{\abovedisplayskip}{5pt}
\setlength{\belowdisplayskip}{5pt}
\begin{split}   
    {\rm Att}(Q,K,V) &= {\rm softmax}(\frac{QK^T}{\sqrt{d_K}})V, \\
    {\rm MA}(F_{Q},F_{K},F_{V}) &= {\rm Concat}(A_1, \dots, A_h)W^O,
\end{split}
\end{equation}
{where $A_i \!= \!{\rm
 Att}(F_{Q}W^Q_i\!,\!F_{K}W^K_i\!,\!F_{V}W^V_i)$; $W^Q_i$,$W^K_i$, and $W^V_i$ are the projection matrics used to project $F_{Q}$, $F_{K}$, and $F_{V}$ to queries $Q$, keys $K$, and values $V$; $h$ is the number of attention functions ${\rm Att}$ performed in parallel; $d_K$ is the dimensionality of $K$; $W^O$ is a matrix used to project the concatenated features. }

With regard to the ${\rm MSA}$ in the Transformer encoder,
${\rm MSA}(T_{n}^{'}) = {\rm MA}(T_{n}^{'},T_{n}^{'},T_{n}^{'})$.
${\rm MSA}$ linearly projects $T_{n}^{'}$ onto $Q,K${,} and $V$ with different linear projections $h$ times; \textcolor{black}{then, each
${\rm Att}$ 
obtains an attention map by the scaled dot-product between $Q$ and $K$ to consider all relations between
each point in $src$, and multiplies this map by $V$ to
aggregate information from the entire $src$.}
{By conducting ${\rm MSA}$ and ${\rm MLP}$, the encoder obtains $T_{n+1}$ based on ${\rm MLP}$ and ${\rm MSA}$ as}
\begin{equation} \label{encoder_T} \small
\setlength{\abovedisplayskip}{5pt}
\setlength{\belowdisplayskip}{5pt}
\begin{split}
    \hat{T} &= {\rm LN}({\rm MSA}(T_{n}^{'})) + T_{n}^{'}), \\
    T_{n+1} &= {\rm LN}({\rm MLP}(\hat{T})) + \hat{T}). \\
\end{split}
\end{equation}

Finally, obtaining all output values $T_n$ of Transformer encoders, with $ n\! =\! 2,\cdots, N_l\!+\!1$, low-order and high-order features are merged by concatenating all features $T_n$ as
\begin{equation} \label{concatenating} \small
\setlength{\abovedisplayskip}{5pt}
\setlength{\belowdisplayskip}{5pt}
    \begin{split}
        F_X = {\rm LN}({\rm ReLU}({\rm Concat}(T_2,T_3,\cdots,T_{N_l+1}))).
    \end{split}
    \end{equation}




\subsection{\vspace{0pt}Point Feature Transformer\vspace{0pt}}
{The features $F_X$ and $F_Y$ extracted by PSE are still independent of each other, which leads to an indistinct similarity matrix. Therefore, to learn the {contextual information of two point clouds and extract discriminative features}, PFT is designed to facilitate deep information interaction.}

{Since the standard Transformer can not directly  learn the relative position between points \cite{vaswani2017attention}, positional encoding is introduced to extract positional information.} 
{To
model the positional information $P_X$ with coordinates $X$ \cite{zhao2021point}}, an efficient neural network is introduced  , which {consists of fully connected layers ${\rm FC}$, rectified linear unit {\rm ReLU} activation, and {\rm sigmoid} activation} :
\begin{equation} \label{PE} \small
\setlength{\abovedisplayskip}{5pt}
\setlength{\belowdisplayskip}{5pt}
    P_X  = {\rm ReLU}({\rm FC}({\rm Sigmoid}({\rm FC}(X)))).
\end{equation}

Subsequently, we sum the positional information $P_X$ and $P_Y$ with the extracted features $F_{X}$ and $F_{Y}$ to obtain features $F _{X}^{'}$ and $F _{Y}^{'}$, respectively.

{To aggregate information from $src$ and $tgt$, a standard Transformer ${\rm \phi}$ is
adopted, which
consists of an encoder (Eq. \ref{encoder_T}) and a decoder. The
Transformer decoder consists of a {multi-head} cross-attention
operation (${\rm MCA}$), in addition to ${\rm MSA}$, ${\rm MLP}$,
and ${\rm LN}$ (Eq. \ref{self-Attention}). Taking ${\rm
\phi}(F_{Y}^{'},F_{X}^{'})$ as an example, the procedure is defined as}
\begin{equation} \label{decoder_T} \small
\setlength{\abovedisplayskip}{5pt} 
\setlength{\belowdisplayskip}{5pt}
\begin{split}
    F_{Xa} &= {\rm LN}({\rm MSA}(F_{X}^{'}) + F_{X}^{'}), \\
    F_{Xa}^{'} &= {\rm LN}({\rm MCA}(F_{Ya},F_{Xa}) + F_{Xa}), \\
    F_{Xo} &= {\rm LN}({\rm MLP}(F_{Xa}^{'}) + F_{Xa}^{'}), \\
\end{split}
\end{equation}
where  ${\rm MCA}(F_{Ya},F_{Xa})={\rm MA}(F_{Xa},F_{Ya},F_{Ya})$; features $F_{Xa}$ are obtained based on ${\rm MSA}$; features $F_{Ya}$ are acquired through the encoder;
 {the attention map  is acquired in ${\rm MCA}$ to establish the associations across points in $X$ and  $Y$}, which enables $F_{Xa}$ to receive information from $F_{Ya}$ and improves the {discrimination}  of $F_{Xa}$.

However, due to the shallow-wide architecture used in previous methods, the associations established for information interaction are limited, leading to low feature richness. In this paper, we instead utilize a deep-narrow architecture to establish comprehensive associations. 

%

Overall, the feature vectors $\Psi  _{X}$ and $\Psi _{Y}$ generated by {the Transformer} are formulated as
\begin{equation} \label{EN} \small
\setlength{\abovedisplayskip}{5pt}
\setlength{\belowdisplayskip}{5pt}
    \begin{split}
        \Psi  _{X} & = F_{X} +  {\rm \phi}(F _{Y}^{'} , F _{X}^{'}),  \\ 
        \Psi _{Y} & = F_{Y} +  {\rm \phi}(F _{X}^{'} , F _{Y}^{'}).
    \end{split}
\end{equation}

{To adaptively recalibrate the channel-wise features in accordance with their contribution in registration, a squeeze-and-excitation (SE) module \cite{hu2018squeeze} is adopted.} The SE module first extracts a channel descriptor $F_{sq}$ by applying average pooling to the input features $F_{in}$, then maps $F_{sq}$ to channel weights $F_{ex}$ by means of a  neural network, and finally rescales $F_{in}$ with  $F_{ex}$ to obtain  rescaled features $F_{c} = F_{ex}F_{in}$.

In summary, by applying the above positional encoding network, Transformer model, and SE module, the feature vectors $\Phi _{X}$ and $\Phi _{Y}$ generated by PFT are defined as
\begin{equation} \label{PFT} \small
\setlength{\abovedisplayskip}{5pt}
\setlength{\belowdisplayskip}{5pt}
    \begin{split}
    \Phi _{X} & = {\rm SE}(F_{X} + {\rm \phi}(F_{Y}+P_Y , F_{X}+P_X)),  \\ 
    \Phi _{Y} & = {\rm SE}(F_{Y} + {\rm \phi}(F_{X}+P_X , F_{Y}+P_Y)).
    \end{split}
\end{equation}


\begin{spacing}{1.1}
\begin{algorithm}[t] 
    \caption{Correspondence Confidence Evaluation}  
    \label{alg:Framwork}  
    \begin{algorithmic}[1] 
        \small
      \Require Point clouds $X \in R^{N \times 3}$, $ Y \in R^{M \times 3}$ and correspondence $M \in R^{N \times 1}$
      \Ensure  
        Confidence $C \in R^{N \times 1}$ of correspondence $M$ 
      \State $S_{src} \leftarrow {\rm KNN}(X, k)$
      \label{code:fram:extract}
      \State $P_{src} \leftarrow {\rm combinations(S_{src},2)}$
      \label{code:fram:trainbase}  
      \State $E_{src} \leftarrow X.{\rm reshape}(N,1,1,3).{\rm repeat}(1,{\frac{(k^2-k)}{2}} , 1 , 1)$ 
      \label{code:fram:add}  
      \State $g_{src} \leftarrow {\rm concatenate}(E_{src}, P_{src})$  
      \label{code:fram:classify}  
      \State $(l_g^{src},l_g^{tgt}) \leftarrow {\rm GetTri}(g_{src} , g_{tgt})$  
      \label{code:fram:select} 
      \State $L_e \leftarrow {{\rm sum}((l_g^{src}-l_g^{tgt})^{**}2, dim=-1)}/{\rm sum}((l_g^{src}+l_g^{tgt})^{**}2, dim=-1)$  
      \label{code:fram:classify}
      \State $E_r \leftarrow {\rm sum}({\rm sqrt}({\rm Mink}(L_e)), dim=-1)$  
      \label{code:fram:classify}
      \State $C \leftarrow {\psi}(2\times {\rm sigmoid}(-\lambda E_r))$
      \label{code:fram:classify}
      \\  
      \Return $C$  
    \end{algorithmic}  
  \end{algorithm} 
\end{spacing}
\subsection{Geometric Matching-Based 
Correspondence Confidence Evaluation}
\begin{figure}[t]
\setlength{\abovecaptionskip}{4pt}
\centering
\includegraphics[width=8.3cm]{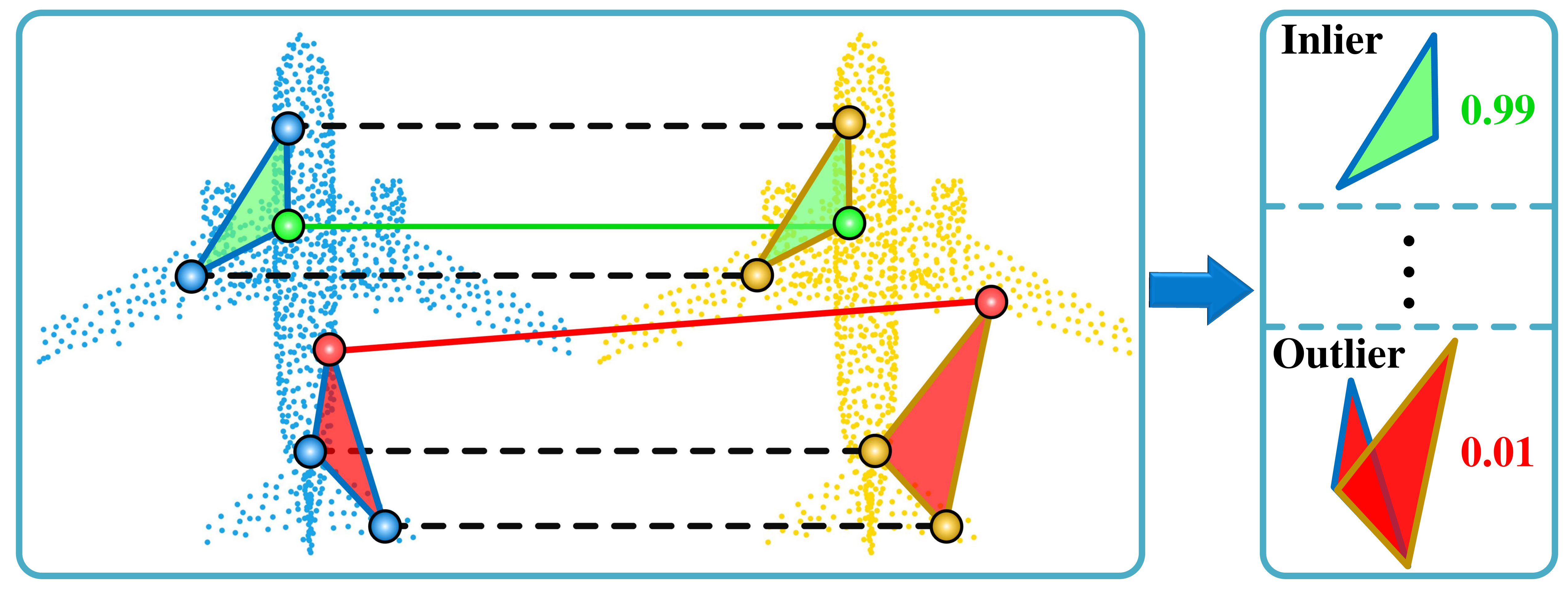}
\caption{\textbf{The demonstration of the  triangulated descriptor.}  When
the correspondence $\{x_{i}, y_{j}\}$ is an inlier,
 the corresponding triangles are similar, and the confidence value $C(x_{i}, y_{j})$ is high.
 Otherwise, $C$ decreases to a small value if $\{x_{i}, y_{j}\}$ is an outlier. }
\label{CCV}\vspace{-12pt}
\end{figure}

Given features $\Phi _{X}$ and $\Phi _{Y}$, a set of putative correspondences $M\{x_i,y_j\}$ are established by finding the most similar features $\{\Phi _{xi}, \Phi _{yj}\}$.
However, there are outliers in partial-to-partial point cloud registration, which significantly reduces the accuracy. Therefore,
the GMCCE module is designed to distinguish between inliers and outliers with a representative descriptor.
To accurately describe {geometric characteristics}, a triangulated descriptor is designed. As shown in Fig. \ref{CCV}, the descriptor employs the side length of triangles to capture geometric characteristics. 
It offers two advantages: 
(1) expressing length and angle simultaneously;
(2) establishing connections between sampled points.

The GMCCE module is presented in Fig. \ref{Overview}(d), and the detailed procedures are shown in Algorithm \ref{alg:Framwork}.  First, ${\rm KNN}$ is used to search for $k_s$ sampled points $S_{src}$ of $x_i$ in $src$, and then, testing groups $g_{src}$ are obtained by combining $S_{src}$ and $x_i$; specifically, each group contains $x_i$ and two points of $S_{src}$. Afterward, testing groups $g_{tgt}$ are acquired by mapping  $g_{src}$ in accordance with the correspondence matrix $M$. Subsequently, the lengths $l_g^{src}$ and $l_g^{tgt}$ of the triangles constructed by $g_{src}$ and $g_{tgt}$, respectively, are calculated. Finally, the overall error $E_r(x_i,y_j)$ is calculated by summing the $k$ smallest errors $L_e$ of each group as
\begin{equation}\label{ERR_equation} \small
\setlength{\abovedisplayskip}{5pt}
\setlength{\belowdisplayskip}{5pt}
\begin{split}
    L_e(g_{src}, g_{tgt}) = &\sqrt{{ \Sigma(l_{g\alpha }^{src} \!-\! l_{g\alpha }^{tgt})^2}/
    { \Sigma(l_{g\alpha }^{src} \!+\! l_{g\alpha }^{tgt})^2}}, \\
    E_r(x_{i}, y_{j})\! = \! \Sigma  {\rm Mink}&([L_e(g_{src}^1,g_{tgt}^1),\!\cdots,\! L_e(g_{src}^P,g_{tgt}^P)]),
\end{split}    
\end{equation}
where $ l_{g\alpha }^{src} $ and $ l_{g\alpha }^{tgt} $ denote the side lengths of
the triangles constructed by $g_{src}$ and $g_{tgt}$, respectively; ${\rm Mink}$ is the operation of taking the $k$ smallest values{;} $P$  equals $\frac{k(k-1)}{2}$. Then, the confidence $C(x_i,y_j)$ is evaluated as
\begin{equation}\label{CCV_equation} \small
\setlength{\abovedisplayskip}{5pt}
\setlength{\belowdisplayskip}{5pt}
    C(x_{i}, y_{j}) = {\rm \psi}(2 \times {\rm sigmoid}(-\lambda E_r(x_{i}, y_{j}))), 
\end{equation}
where $\lambda$ is the parameter to adjust the sharpness of the confidence evaluation; $\psi$ is the filter
 to filter out correspondences with confidence smaller than $\tau $.

\subsection{Loss Function}
The overall loss function to train our DIT consists of three terms: a transformation loss $L_t$, a cycle
consistency loss $L_c$, and a discrimination loss $L_d$.
By combining these terms {and introducing coefficients $\alpha$ and $\beta$ to  adjust the contribution of each loss term}, the final loss function is constructed and defined as 
\begin{equation} \small
\setlength{\abovedisplayskip}{5pt}
\setlength{\belowdisplayskip}{5pt}
\label{loss_function}
        L = L_t + \alpha L_c + \beta L_d.
\end{equation}

\textbf{Transformation loss}: $L_t$  measures the error between the predicted motion $R_{XY}$, $t_{XY}$ and ground-truth motion $R_{XY}^*, t_{XY}^*$ from $X$ to $Y$ as
\begin{equation} \small
\setlength{\abovedisplayskip}{5pt}
\setlength{\belowdisplayskip}{5pt}
    L_t =\| R_{XY}^{T}R_{XY}^{*} - I \Vert^2 + \| t_{XY}^{T}t_{XY}^{*} - I \Vert^2. 
\end{equation}

\textbf{Cycle consistency loss}: $L_c$  measures the consistency between the predicted motion $R_{XY}$, $t_{XY}$ from $X$ to $Y$ and $R_{YX}$, $t_{YX}$ from $Y$ to $X$ as
\begin{equation} \small
\setlength{\abovedisplayskip}{5pt}
\setlength{\belowdisplayskip}{5pt}
    L_c = \| R_{XY}^{T}R_{YX} - I \Vert^2 + \| t_{XY}-t_{YX} \Vert^2. 
\end{equation}

\textbf{Discrimination loss}: $L_d$ measures the discriminative power of  extracted features and the accuracy of  established correspondences as
\begin{equation} \small
\setlength{\abovedisplayskip}{5pt}
\setlength{\belowdisplayskip}{5pt}
    \begin{split}
        L_d &= -\frac{1}{\|M \Vert } \sum_{((x_i,y_j)\in M)}[C(i,j) \times \ln S(i,j)  \\
    &+(1-C(i,j))\times\ln (1-S(i,j))],  
    \end{split}
\end{equation}
where $C(i,j)=1$ if the correspondence $\{x_i, y_j\}$ is an inlier; otherwise, $C(i,j)=0$.
$S(i,j)$ denotes the similarity between the feature vectors $\Phi _{xi}$ and $\Phi _{yj}$.


\section{Experimental Results}
\subsection{{Experimental Setup}}\label{dataset}
  \textbf{Dataset}: The proposed algorithm and  baseline methods are evaluated on ModelNet40 \cite{wu20153d}. This dataset includes 12,311 meshed computer-aided design {(CAD)} models in 40 categories, of which $80\%$ are designated for training and the remaining $20\%$ are designated for testing. We randomly sample 1,024 points on the surface of a model as $src$ and rescale the points to a unit sphere. An initial rigid transformation is randomly sampled from the following  intervals: the rotation along each axis  in $[ 0,45^{\circ} ]$, and the translation along each axis 
 in $[ -0.5,0.5 ]$.  This initial transformation is then applied to $src$ to obtain $tgt$. 
 
 \textbf{ {Implementation Details}}:
 Each LFI layer concatenates the feature vectors from a neighborhood of $k = 20$ and
 outputs features with $64$ dimensions. The ${\rm MA}$ modules in the PSE and PFT networks each have
 $4$ heads. In the GMCCE module, the parameters $\lambda = 30$, and $k_s=10$  are obtained by grid search.   The network is trained using the Adam \cite{kingma2014adam} optimizer with an initial learning rate of 3e-5. 

\textbf{{Comparison methods}}:
DIT is compared against the representative model-based methods:
ICP \cite{besl1992method}, FGR \cite{zhou2016fast}, and
 FPFH \cite{rusu2009fast} + RANSAC \cite{fischler1981random} and recent learning-based methods:
PointNetLK (PNetLK)  \cite{aoki2019pointnetlk}, DCP \cite{wang2019deep}, DeepGMR \cite{yuan2020deepgmr}, IDAM \cite{li2020iterative},
Reagent \cite{bauer2021reagent}, PNetLK\_R \cite{li2021pointnetlk}, and RGM \cite{fu2021robust}. All experiments are evaluated on an Intel i7-10700 CPU and an RTX 3090 graphics card.
ICP , FGR, and FPFH are implemented with the Intel Open3d library \cite{zhou2018open3d}.
For the other methods, we reproduced the open source code provided by the published papers with the same settings and hyperparameters. 

\textbf{Evaluation metrics}: We evaluate the performance of each point cloud registration method using the root mean squared error (RMSE) and the mean absolute error (MAE). All metrics related to rotation are expressed in units of degrees. 
\textcolor{black}{Comparisons are tested for three scenarios: (1) clean point clouds, (2) low noise partial-to-partial point clouds,  (3) high noise partial-to-partial point clouds. }

\begin{figure*}
    \begin{center}
    \includegraphics[width=17.5cm]{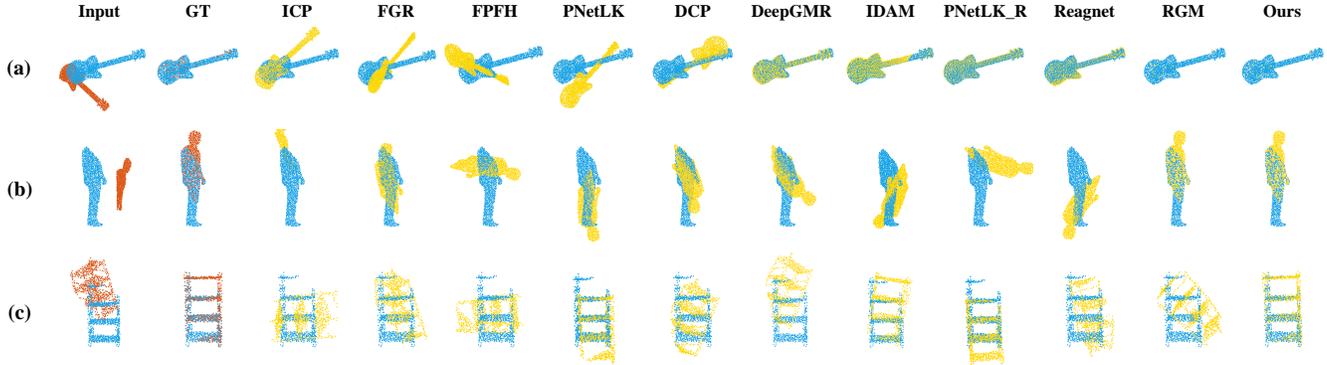}
    \end{center}
    \setlength{\abovecaptionskip}{-0.15cm}
       \caption{\textbf{Qualitative registration results on ModelNet40.} (a) Clean point clouds. (b) Low noise partial-to-partial point clouds. (c) High noise partial-to-partial point clouds.}
    \label{results}
    \vspace{-10pt}
\end{figure*}

\subsection{Matching and Registration Performance}

\textbf{Clean point clouds}:
We first evaluate the performance on clean point clouds. Qualitative  results are shown in Fig.  \ref{results}(a) and quantitative
comparisons are summarized in Table  \ref{clean point clouds}. {Our method} achieves the best performance. Compared with the second-best {{RGM}}, our method reduces the rotation and translation errors significantly.
{The experimental results show that the deep-narrow architecture of the Transformer and the introduction of positional encoding sharpen the mapping for point cloud alignment}. Furthermore, the results verify the ability of {the} proposed method to identify the structure of the point cloud and the effectiveness in distinguishing between inliers and outliers. 

\setlength{\tabcolsep}{0.4mm}
{
    \begin{table}[t]
    \setlength{\abovecaptionskip}{4pt}
        \small
        \center
        \caption{\textbf{Performance on clean point clouds.} The three best  results are  highlighted in \textbf{\textcolor{red}{red}}, \textbf{\textcolor{green}{green}},  \textbf{\textcolor{blue}{blue}}.}
        \label{clean point clouds}
        \begin{tabular}{l@{\hspace{10pt}}c@{\hspace{5pt}}c@{\hspace{5pt}}c@{\hspace{5pt}}c@{\hspace{5pt}}c} 
            \toprule[1.5pt]
            {\small Method}  & {\small Reference}  & {\small ${\rm R_{RMSE}}$} & {\small ${\rm R_{MAE}}$} &{\small ${\rm t_{RMSE}}$} &{\small ${\rm t_{MAE}}$}   \\
            \midrule[1pt]
            ICP \cite{besl1992method}     & SPIE 1992        &25.09      &14.15   &0.157    &0.106   \\
            FGR \cite{zhou2016fast}    & ECCV 2016          &10.63&2.335&0.014&0.0045          \\
            FPFH \cite{rusu2009fast}&ICRA 2009&15.40&2.643&0.048&0.0090\\
            PNetLK \cite{aoki2019pointnetlk}&CVPR 2019&12.02&4.954&0.0064&0.0038 \\
            DCP\_V2 \cite{wang2019deep}&CVPR 2019&3.242&2.076&0.0024&0.0015\\
            DeepGMR \cite{yuan2020deepgmr}&ECCV 2020&\textbf{\textcolor{blue}{0.023}}&\textbf{\textcolor{blue}{0.016}}&\textbf{\textcolor{blue}{3e-5}}&\textbf{\textcolor{blue}{2e-5}} \\
            IDAM \cite{li2020iterative}&ECCV 2020&1.59&1.109&0.0259&0.018 \\
            PNetLK\_R \cite{li2021pointnetlk}&CVPR 2021& 1.385&0.120&0.0085&0.0006\\
            Reagent \cite{bauer2021reagent}&CVPR 2021&1.073&0.939&0.0023&0.0020\\
            RGM \cite{fu2021robust}&CVPR 2021&\textbf{\textcolor{green}{1.8e-4}}&\textbf{\textcolor{green}{1.2e-5}}&\textbf{\textcolor{green}{2.7e-6}}&\textbf{\textcolor{green}{1.6e-7}}\\
            \textbf{Ours}&-&\textbf{\textcolor{red}{2.3e-6}}&\textbf{\textcolor{red}{1.5e-6}}&\textbf{\textcolor{red}{1.7e-8}}&\textbf{\textcolor{red}{1.1e-8}}\\
            \bottomrule[1.5pt]
        \end{tabular}\vspace{-6pt}
    \end{table}
}

\textbf{Low noise partial-to-partial point clouds}:
Partial-to-partial registration is much more challenging due to the existence of outliers and the difficulty of extracting contextual information. 
Following the similar operation of generating partial-to-partial point clouds  in PRNet \cite{wang2019prnet}, we remove 200 points from each \textit{src} and \textit{tgt} to obtain a point cloud pair with an overlap rate of approximately 60\% (IoU). Then, Gaussian noise sampled from $\mathcal{N}  (0,0.001)$  and clipped to $[-0.001,0.001]$  is added to each point.
The results on low noise partial-to-partial registration are shown in Fig.  \ref{results} and Table  \ref{partial low}. 
{{Our method}} clearly outperforms the other methods; specifically, the rotation and translation errors are reduced obviously compared with  {{RGM}}. {Due to the lack of information aggregation}, the accuracy of DeepGMR and PNetLK\_R is much lower than that on clean point clouds. 
The experimental results verify the importance of aggregating information from the two point clouds and show that DIT can extract contextual information by means of deep information interaction, enabling DIT to precisely identify common structures.

\setlength{\tabcolsep}{0.7mm}
{
    \begin{table}[t]
    \setlength{\abovecaptionskip}{4pt}
        \small
        \center
        \caption{\textbf{Performance on low noise partial-to-partial point clouds.} The three best  results are  highlighted in \textbf{\textcolor{red}{red}}, \textbf{\textcolor{green}{green}},  \textbf{\textcolor{blue}{blue}}.}
        \label{partial low}
        \begin{tabular}{l@{\hspace{10pt}}c@{\hspace{5pt}}c@{\hspace{5pt}}c@{\hspace{5pt}}c@{\hspace{5pt}}c}
            \toprule[1.5pt]
            {\small Method}  & {\small Reference}  & {\small ${\rm R_{RMSE}}$} & {\small ${\rm R_{MAE}}$} &{\small ${\rm t_{RMSE}}$} &{\small ${\rm t_{MAE}}$}   \\
            \midrule[1pt]
            ICP \cite{besl1992method}   & SPIE 1992          & 20.12            &11.24            &0.13            &0.092             \\
            FGR \cite{zhou2016fast}       & ECCV 2016     &21.4            &7.03            &0.063            &0.025                  \\
            FPFH \cite{rusu2009fast} &ICRA 2009&29.97            &8.38            &0.083            &0.023            \\
            PNetLK \cite{aoki2019pointnetlk}&CVPR 2019&18.10            &12.38            &0.131            &0.101             \\
            DCP\_V2 \cite{wang2019deep}&CVPR 2019&\textbf{{\textcolor{blue}{4.43}}}            &\textbf{\textcolor{blue}{2.92}}            &\textbf{\textcolor{blue}{0.029} }           &\textbf{\textcolor{blue}{0.022}}            \\
            DeepGMR \cite{yuan2020deepgmr}&ECCV 2020&7.15            &4.84            &0.13            &0.107             \\
            IDAM \cite{li2020iterative}&ECCV 2020&{\textcolor{black}{14.44}}&\textcolor{black}{8.54}&{\textcolor{black}{0.10}}&{\textcolor{black}{0.07}} \\
            PNetLK\_R \cite{li2021pointnetlk}&CVPR 2021 &7.37&6.32&0.062&0.053\\
            Reagent \cite{bauer2021reagent}&CVPR 2021&9.37            &8.22            &0.055            &0.043            \\
            RGM \cite{fu2021robust} &CVPR 2021&\textbf{\textcolor{green}{0.741}}           &\textbf{\textcolor{green}{0.099 } }          &\textbf{\textcolor{green}{2.4e-3}}            &\textbf{\textcolor{green}{8.1e-4} }         \\
            \textbf{Ours}&-&\textbf{\textcolor{red}{0.014 }}           &\textbf{\textcolor{red}{0.010  }}&\textbf{\textcolor{red}{6.7e-5 }} &\textbf{\textcolor{red}{5.3e-5  }}\\
            \bottomrule[1.5pt]
        \end{tabular}\vspace{-6pt}
    \end{table}
}

\textbf{High noise partial-to-partial point clouds}:
To evaluate the robustness against high noise in  partial-to-partial registration tasks, similar to the operation in PRNet \cite{wang2019prnet},
Gaussian noise independently sampled from $\mathcal{N} (0,0.01)$ and clipped to $[-0.05,0.05]$ is added to each point. The other experimental settings are the same as in the
low noise  experiment.
The  results  on the high  noise partial-to-partial registration  are shown in Fig.  \ref{results}(c) and  Table  \ref{partial high}.
{{Our method}} still outperforms  other methods; specifically, our method improves the rotation and translation accuracy by {32\%--65\%} compared with {{RGM}}. We also note that the accuracy of DCP is reduced by approximately 60\% compared with its accuracy in the low noise case. The results reveal that DIT can model global relations, thereby achieving superior robustness against high noise.

\subsection{{Accuracy and Generalization} Analysis}
To compare the accuracy and generalization, we present the success ratios of all methods  in Fig.  \ref{Accuracy analysis}, where the success ratio is defined as the ratio of error ($R_{error},t_{error}$) less than the threshold ($R_{thres},t_{thres}$). Our method achieves the best performance in all three different settings. With the tightening of  convergence thresholds, \textcolor{black}{ our method always achieves the highest accuracy and generalization}. Fig.  \ref{Accuracy analysis}(a) shows the success ratios on clean point clouds. Due to the strict convergence thresholds, several methods did not converge, only {RGM} and PNetLK\_R are competitive with our method. Our method is clearly the only approach that achieves both the fastest convergence and a success rate of {100\%}. Specifically, PNetLK\_R reaches a {98\%} success rate, but the error is very large when the matching fails, leading to high MAE and RMSE values in  Table \ref{clean point clouds}.

\setlength{\tabcolsep}{0.4mm}
{
    \begin{table}[t]
    \setlength{\abovecaptionskip}{4pt}
        \small
        \center
        \caption{\textbf{Performance on high noise partial-to-partial point clouds.} The three best  results are  highlighted in \textbf{\textcolor{red}{red}}, \textbf{\textcolor{green}{green}},  \textbf{\textcolor{blue}{blue}}.}
        \label{partial high}
        \begin{tabular}{l@{\hspace{10pt}}c@{\hspace{5pt}}c@{\hspace{5pt}}c@{\hspace{5pt}}c@{\hspace{5pt}}c}
            \toprule[1.5pt]
            {\small Method}  & {\small Reference}  & {\small ${\rm R_{RMSE}}$} & {\small ${\rm R_{MAE}}$} &{\small ${\rm t_{RMSE}}$} &{\small ${\rm t_{MAE}}$}   \\
            \midrule[1pt]
            ICP \cite{besl1992method}       & SPIE 1992      & 20.05            &11.33            &0.13            &0.09            \\
            FGR \cite{zhou2016fast}       &ECCV 2016     &47.58            &27.27            &0.126            &0.088                  \\
            FPFH \cite{rusu2009fast} &ICRA 2009&55.47            &28.9            &0.165            &0.087            \\
            PNetLK \cite{aoki2019pointnetlk}&CVPR 2019&18.33            &12.17            &0.14            &0.108             \\
            DCP\_V2 \cite{wang2019deep}&CVPR 2019&12.29            &7.84            &0.097            &0.078            \\
            DeepGMR \cite{yuan2020deepgmr}&ECCV 2020&\textbf{\textcolor{blue}{8.957}}            &\textbf{\textcolor{blue}{6.243} }           &0.154            &0.128             \\
            IDAM \cite{li2020iterative}&ECCV 2020&{\textcolor{black}{18.91}}&{\textcolor{black}{11.78}}&{\textcolor{black}{0.093}}&{\textcolor{black}{0.067}} \\
            PNetLK\_R \cite{li2021pointnetlk} &CVPR 2021&9.837&7.626&0.093&0.075\\
            Reagent \cite{bauer2021reagent}&CVPR 2021&11.85            &10.47            &\textbf{\textcolor{blue}{0.063}}            &\textbf{\textcolor{blue}{0.05}}            \\
            RGM \cite{fu2021robust}&CVPR 2021&\textbf{\textcolor{green}{2.068}}            &\textbf{\textcolor{green}{0.633}}            &\textbf{\textcolor{green}{0.016}}            &\textbf{\textcolor{green}{0.0061}}            \\
            \textbf{Ours}&-&\textbf{\textcolor{red}{1.412            }}&\textbf{\textcolor{red}{0.357            }}&\textbf{\textcolor{red}{0.009            }}&\textbf{\textcolor{red}{0.0021            }}\\
            \bottomrule[1.5pt]
        \end{tabular}\vspace{-6pt}
    \end{table}
}

\begin{figure}[t]
\setlength{\abovecaptionskip}{5pt}
\centering
\includegraphics[width=8.3cm]{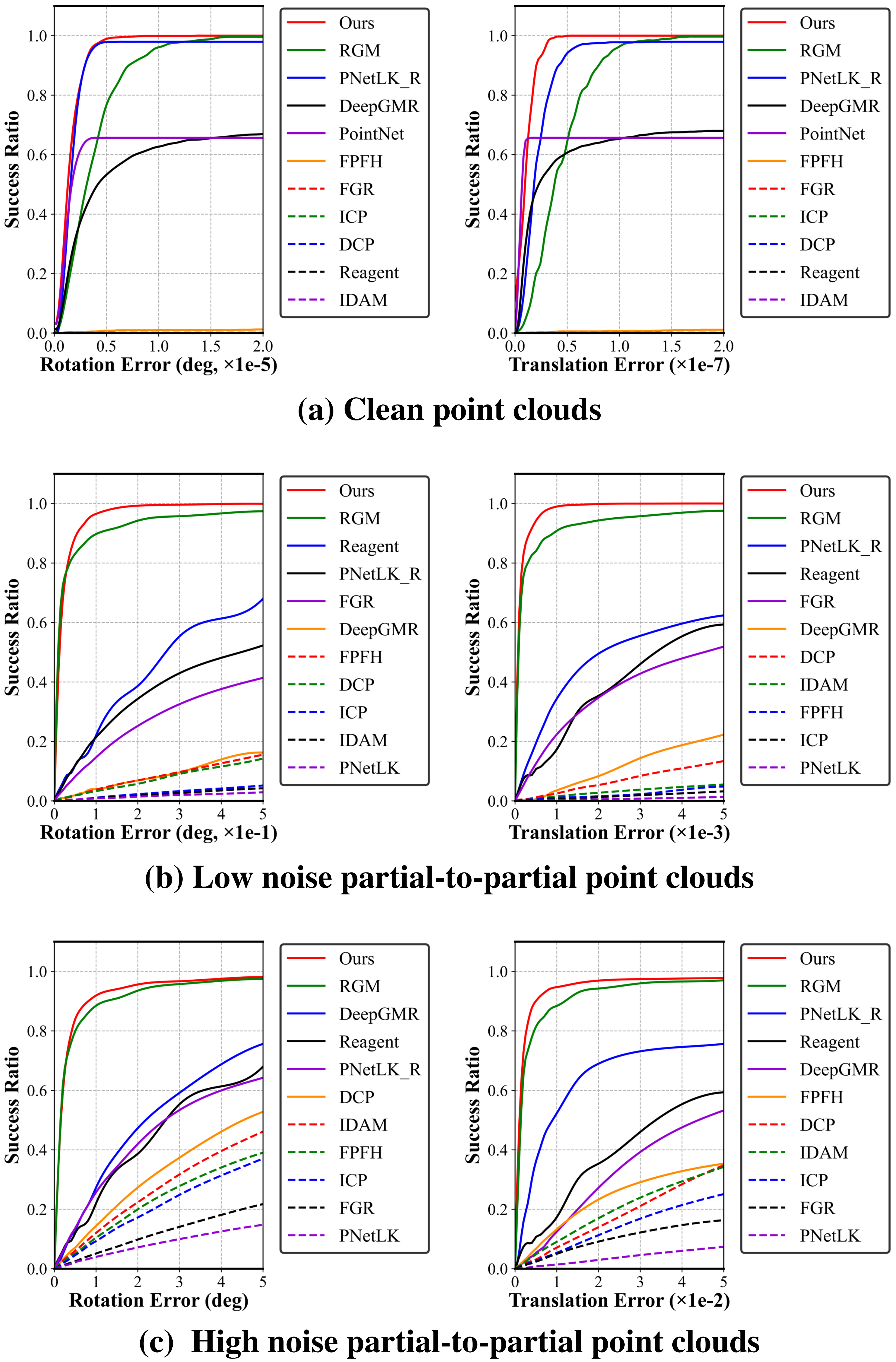}
\caption{\textbf{{Accuracy and generalization analysis}.}  The three best  results are shown as solid lines in \textbf{\textcolor{red}{red}}, \textbf{\textcolor{green}{green}}, and \textbf{\textcolor{blue}{blue}}. }
\label{Accuracy analysis}\vspace{-13pt}
\end{figure}

As shown in Fig. \ref{Accuracy analysis}(b), {our method} is the only approach with an ultimate success ratio of {99\%} in both rotation and translation. Only {RGM} reaches {94\%}, whereas all the other methods  are less than {80\%}. Compared with the previous experiment, PNetLK\_R performs poorly in this partial-to-partial task.
\textcolor{black}{The results reveal that the deep-narrow architecture and positional encoding can improve the accuracy on partial-to-partial point clouds}. Fig.  \ref{Accuracy analysis}(c) shows the success ratios on the more challenging high noise partial-to-partial point clouds.
{Our method} still surpasses the other methods, achieving a {98\%} success rate in both rotation and translation.
The results demonstrate that modeling global relations  strengthens the robustness to high noise, and our full Transformer framework achieves high accuracy and generalization in various registration tasks. 

\subsection{Ablation Studies}
\begin{figure}[t]
\setlength{\abovecaptionskip}{5pt}
\setlength{\belowcaptionskip}{-0.1cm}
    \centering
    \includegraphics[width=8.3cm]{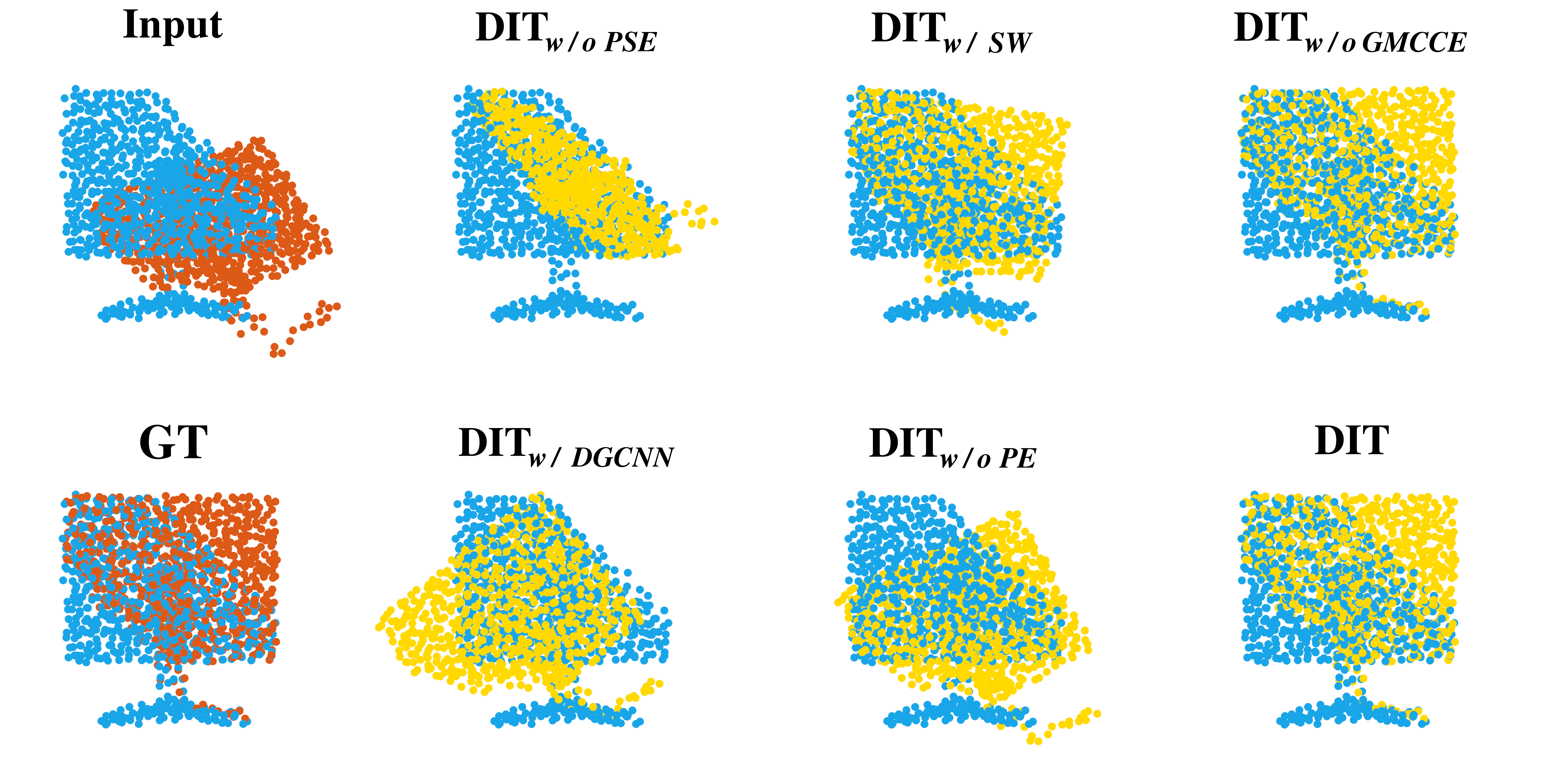}
    \caption{{Qualitative ablation results on ModelNet40.}}
    \label{Ablation results}
    \end{figure}

To analyze the effectiveness of the proposed three key components (PSE, PFT, and GMCCE), we present ablation studies  by comparing the performance of five variants on the high noise setting. 
The results of the five variants are shown in Table  \ref{ablation} and Fig.  \ref{Ablation results}.


\textbf{PSE:}  ${\rm DIT}_{w/o \; PSE}$ is designed to exclude the PSE network, the success ratio drops by 93.5\%. ${\rm DIT}_{w/ \; DGCNN}$ substitutes the PSE module with the DGCNN module \cite{phan2018dgcnn}, and the success ratio drops by {87.2\%}. The results signify  the effectiveness of the 
PSE network in modeling global relations and identifying structural characteristics.  

\textbf{PFT:} 
${\rm DIT}_{w/ \; SW}$ is designed to use a shallow-wide architecture. With this variant, the accuracy declines by 55\%--80\%, which shows that the deep-narrow architecture {establishes more comprehensive associations and facilitates the deep information interaction.}
${\rm DIT}_{w/o \; PE}$ is designed to exclude the positional encoding network. The results show that the success ratio declines by {39.4\%},
demonstrating that the positional encoding network enables the Transformer to directly learn the relative position {between} points, {enhancing} the robustness to noise.

\textbf{GMCCE:}   ${\rm DIT}_{w/o \; GMCCE}$ is designed to exclude the GMCCE module.
The rotation and translation accuracy drops by {40\% -- 74\%},
indicating that GMCCE significantly improves the registration accuracy with the advantage of distinguishing between inliers and outliers.

   \begin{table}
        \normalfont
        \center
        \caption{ \textbf{Ablation {results} concerning the effect of PSE, PFT{,} and GMCCE module.}  SR denotes the success rate ($R_{thres}=1,t_{thres}=0.01$).}
        \label{ablation}
        \begin{tabular}{l@{\hspace{8pt}}c@{\hspace{5pt}}c@{\hspace{5pt}}c@{\hspace{5pt}}c@{\hspace{5pt}}c}
            \toprule[1.5pt]
            {\small Method}  & {\small ${\rm R_{RMSE}}$} & {\small ${\rm R_{MAE}}$} &{\small ${\rm t_{RMSE}}$} &{\small ${\rm t_{MAE}}$}  & {\small ${\rm SR}$}  \\
            \midrule[1pt]
            ${\rm DIT}_{w/o \; PSE}$           &41.84            &28.74             &0.247             & 0.197    &  1.2\%             \\
            ${\rm DIT}_{w/ \; DGCNN}$ &18.07            &12.19             &0.069            &0.049  & 7.5\%           \\
            ${\rm DIT}_{w/ \; SW }$  &7.06           & 1.618          &   0.020         &0.009         & 71\%            \\
            ${\rm DIT}_{w/o \; PE}$           &18.02            &6.013             &0.091             &0.038 &55.3\%                 \\
            ${\rm DIT}_{w/o \; GMCCE}$ & 2.36             & 1.04            &0.016            &0.008      & 74.4\%        \\
            ${\rm DIT}$ &\textbf{1.41            }&\textbf{0.357            }&\textbf{0.009            }&\textbf{0.0021}  &\textbf{94.7\%}       \\
            \hline
        \end{tabular}\vspace{-8pt}
    \end{table}

\vspace{-3pt}
\subsection{Limitations}
 The experiments have demonstrated the high accuracy and robustness of the DIT on ModelNet40. Since the computational complexity of the attention mechanism is quadratic of the point cloud scale, the application to large-scale point clouds requires downsampling or voxelization pre-processing, such operations will reduce the registration accuracy and robustness. Therefore, accurately performing large-scale point cloud registration is a general challenge faced by  Transformer-based methods. Several recent methods \cite{choromanski2020rethinking
,wang2020linformer} explored how to improve the efficiency of the attention model, but also caused a certain level of performance degradation.
\vspace{-3pt}

\section{Conclusion}
{In this work, we explore and propose a novel full Transformer framework DIT for point cloud registration. The DIT effectively models global relations, enhances feature richness, and removes outliers, overcoming the limitations of previous Transformer-based methods.
In DIT, PSE is utilized to model dependencies in the entire point cloud and identify the characteristic of neighboring structures, enhancing the robustness to noise. {Subsequently,} PFT is proposed to improve the discrimination of extracted features  by facilitating deep information interaction. Moreover, GMCCE is leveraged to enable accurate alignments by detecting inliers based on geometric consistency.}
Extensive experiments have been conducted on ModelNet40, exhibiting that our method outperforms previous methods in terms of accuracy, generalization, and robustness. The results demonstrate the potential of the full Transformer framework in point cloud registration tasks.  In the future, the application to large scale point clouds will be further investigated. 

 
{\small
\bibliographystyle{ieee_fullname}
\bibliography{egbib}

\begin{thebibliography}{10}\itemsep=-1pt

\bibitem{aoki2019pointnetlk}
Yasuhiro Aoki, Hunter Goforth, Rangaprasad~Arun Srivatsan, and Simon Lucey.
\newblock Pointnetlk: Robust \& efficient point cloud registration using
  pointnet.
\newblock In {\em Proceedings of the IEEE/CVF Conference on Computer Vision and
  Pattern Recognition}, pages 7163--7172, 2019.

\bibitem{ba2016layer}
Jimmy~Lei Ba, Jamie~Ryan Kiros, and Geoffrey~E Hinton.
\newblock Layer normalization.
\newblock {\em arXiv preprint arXiv:1607.06450}, 2016.

\bibitem{bauer2021reagent}
Dominik Bauer, Timothy Patten, and Markus Vincze.
\newblock Reagent: Point cloud registration using imitation and reinforcement
  learning.
\newblock In {\em Proceedings of the IEEE/CVF Conference on Computer Vision and
  Pattern Recognition}, pages 14586--14594, 2021.

\bibitem{besl1992method}
Paul~J Besl and Neil~D McKay.
\newblock Method for registration of 3-d shapes.
\newblock In {\em Sensor fusion IV: control paradigms and data structures},
  volume 1611, pages 586--606. International Society for Optics and Photonics,
  1992.

\bibitem{bouaziz2013sparse}
Sofien Bouaziz, Andrea Tagliasacchi, and Mark Pauly.
\newblock Sparse iterative closest point.
\newblock In {\em Computer graphics forum}, volume~32, pages 113--123. Wiley
  Online Library, 2013.

\bibitem{brown2020language}
Tom~B Brown, Benjamin Mann, Nick Ryder, Melanie Subbiah, Jared Kaplan, Prafulla
  Dhariwal, Arvind Neelakantan, Pranav Shyam, Girish Sastry, Amanda Askell,
  et~al.
\newblock Language models are few-shot learners.
\newblock {\em arXiv preprint arXiv:2005.14165}, 2020.

\bibitem{campbell2016gogma}
Dylan Campbell and Lars Petersson.
\newblock Gogma: Globally-optimal gaussian mixture alignment.
\newblock In {\em Proceedings of the IEEE conference on computer vision and
  pattern recognition}, pages 5685--5694, 2016.

\bibitem{campbell2019alignment}
Dylan Campbell, Lars Petersson, Laurent Kneip, Hongdong Li, and Stephen Gould.
\newblock The alignment of the spheres: Globally-optimal spherical mixture
  alignment for camera pose estimation.
\newblock In {\em Proceedings of the IEEE/CVF Conference on Computer Vision and
  Pattern Recognition}, pages 11796--11806, 2019.

\bibitem{carion2020end}
Nicolas Carion, Francisco Massa, Gabriel Synnaeve, Nicolas Usunier, Alexander
  Kirillov, and Sergey Zagoruyko.
\newblock End-to-end object detection with transformers.
\newblock In {\em European Conference on Computer Vision}, pages 213--229.
  Springer, 2020.

\bibitem{chetverikov2005robust}
Dmitry Chetverikov, Dmitry Stepanov, and Pavel Krsek.
\newblock Robust euclidean alignment of 3d point sets: the trimmed iterative
  closest point algorithm.
\newblock {\em Image and vision computing}, 23(3):299--309, 2005.

\bibitem{choromanski2020rethinking}
Krzysztof Choromanski, Valerii Likhosherstov, David Dohan, Xingyou Song,
  Andreea Gane, Tamas Sarlos, Peter Hawkins, Jared Davis, Afroz Mohiuddin,
  Lukasz Kaiser, et~al.
\newblock Rethinking attention with performers.
\newblock {\em arXiv preprint arXiv:2009.14794}, 2020.

\bibitem{choy2020deep}
Christopher Choy, Wei Dong, and Vladlen Koltun.
\newblock Deep global registration.
\newblock In {\em Proceedings of the IEEE/CVF conference on computer vision and
  pattern recognition}, pages 2514--2523, 2020.

\bibitem{choy2019fully}
Christopher Choy, Jaesik Park, and Vladlen Koltun.
\newblock Fully convolutional geometric features.
\newblock In {\em Proceedings of the IEEE/CVF International Conference on
  Computer Vision}, pages 8958--8966, 2019.

\bibitem{deschaud2018imls}
Jean-Emmanuel Deschaud.
\newblock Imls-slam: scan-to-model matching based on 3d data.
\newblock In {\em 2018 IEEE International Conference on Robotics and Automation
  (ICRA)}, pages 2480--2485. IEEE, 2018.

\bibitem{devlin2018bert}
Jacob Devlin, Ming-Wei Chang, Kenton Lee, and Kristina Toutanova.
\newblock Bert: Pre-training of deep bidirectional transformers for language
  understanding.
\newblock {\em arXiv preprint arXiv:1810.04805}, 2018.

\bibitem{dosovitskiy2020image}
Alexey Dosovitskiy, Lucas Beyer, Alexander Kolesnikov, Dirk Weissenborn,
  Xiaohua Zhai, Thomas Unterthiner, Mostafa Dehghani, Matthias Minderer, Georg
  Heigold, Sylvain Gelly, et~al.
\newblock An image is worth 16x16 words: Transformers for image recognition at
  scale.
\newblock {\em arXiv preprint arXiv:2010.11929}, 2020.

\bibitem{eckart2018hgmr}
Benjamin Eckart, Kihwan Kim, and Jan Kautz.
\newblock Hgmr: Hierarchical gaussian mixtures for adaptive 3d registration.
\newblock In {\em Proceedings of the European Conference on Computer Vision
  (ECCV)}, pages 705--721, 2018.

\bibitem{fischler1981random}
Martin~A Fischler and Robert~C Bolles.
\newblock Random sample consensus: a paradigm for model fitting with
  applications to image analysis and automated cartography.
\newblock {\em Communications of the ACM}, 24(6):381--395, 1981.

\bibitem{fu2021robust}
Kexue Fu, Shaolei Liu, Xiaoyuan Luo, and Manning Wang.
\newblock Robust point cloud registration framework based on deep graph
  matching.
\newblock In {\em Proceedings of the IEEE/CVF Conference on Computer Vision and
  Pattern Recognition}, pages 8893--8902, 2021.

\bibitem{gojcic2020learning}
Zan Gojcic, Caifa Zhou, Jan~D Wegner, Leonidas~J Guibas, and Tolga Birdal.
\newblock Learning multiview 3d point cloud registration.
\newblock In {\em Proceedings of the IEEE/CVF conference on computer vision and
  pattern recognition}, pages 1759--1769, 2020.

\bibitem{han2019real}
Lei Han, Lan Xu, Dmytro Bobkov, Eckehard Steinbach, and Lu Fang.
\newblock Real-time global registration for globally consistent rgb-d slam.
\newblock {\em IEEE Transactions on Robotics}, 35(2):498--508, 2019.

\bibitem{hertz2020pointgmm}
Amir Hertz, Rana Hanocka, Raja Giryes, and Daniel Cohen-Or.
\newblock Pointgmm: A neural gmm network for point clouds.
\newblock In {\em Proceedings of the IEEE/CVF Conference on Computer Vision and
  Pattern Recognition}, pages 12054--12063, 2020.

\bibitem{hu2018squeeze}
Jie Hu, Li Shen, and Gang Sun.
\newblock Squeeze-and-excitation networks.
\newblock In {\em Proceedings of the IEEE conference on computer vision and
  pattern recognition}, pages 7132--7141, 2018.

\bibitem{huang2019ccnet}
Zilong Huang, Xinggang Wang, Lichao Huang, Chang Huang, Yunchao Wei, and Wenyu
  Liu.
\newblock Ccnet: Criss-cross attention for semantic segmentation.
\newblock In {\em Proceedings of the IEEE/CVF International Conference on
  Computer Vision}, pages 603--612, 2019.

\bibitem{izatt2020globally}
Gregory Izatt, Hongkai Dai, and Russ Tedrake.
\newblock Globally optimal object pose estimation in point clouds with
  mixed-integer programming.
\newblock In {\em Robotics Research}, pages 695--710. Springer, 2020.

\bibitem{jian2010robust}
Bing Jian and Baba~C Vemuri.
\newblock Robust point set registration using gaussian mixture models.
\newblock {\em IEEE transactions on pattern analysis and machine intelligence},
  33(8):1633--1645, 2010.

\bibitem{kingma2014adam}
Diederik~P Kingma and Jimmy Ba.
\newblock Adam: A method for stochastic optimization.
\newblock {\em arXiv preprint arXiv:1412.6980}, 2014.

\bibitem{koide2021voxelized}
Kenji Koide, Masashi Yokozuka, Shuji Oishi, and Atsuhiko Banno.
\newblock Voxelized gicp for fast and accurate 3d point cloud registration.
\newblock In {\em 2021 IEEE International Conference on Robotics and Automation
  (ICRA)}, pages 11054--11059. IEEE, 2021.

\bibitem{krizhevsky2017imagenet}
Alex Krizhevsky, Ilya Sutskever, and Geoffrey~E Hinton.
\newblock Imagenet classification with deep convolutional neural networks.
\newblock {\em Communications of the ACM}, 60(6):84--90, 2017.

\bibitem{li2020iterative}
Jiahao Li, Changhao Zhang, Ziyao Xu, Hangning Zhou, and Chi Zhang.
\newblock Iterative distance-aware similarity matrix convolution with
  mutual-supervised point elimination for efficient point cloud registration.
\newblock In {\em Computer Vision--ECCV 2020: 16th European Conference,
  Glasgow, UK, August 23--28, 2020, Proceedings, Part XXIV 16}, pages 378--394.
  Springer, 2020.

\bibitem{li2021pointnetlk}
Xueqian Li, Jhony~Kaesemodel Pontes, and Simon Lucey.
\newblock Pointnetlk revisited.
\newblock In {\em Proceedings of the IEEE/CVF Conference on Computer Vision and
  Pattern Recognition}, pages 12763--12772, 2021.

\bibitem{lucas1981iterative}
Bruce~D Lucas, Takeo Kanade, et~al.
\newblock An iterative image registration technique with an application to
  stereo vision.
\newblock Vancouver, British Columbia, 1981.

\bibitem{maron2016point}
Haggai Maron, Nadav Dym, Itay Kezurer, Shahar Kovalsky, and Yaron Lipman.
\newblock Point registration via efficient convex relaxation.
\newblock {\em ACM Transactions on Graphics (TOG)}, 35(4):1--12, 2016.

\bibitem{min2021geometry}
Taewon Min, Eunseok Kim, and Inwook Shim.
\newblock Geometry guided network for point cloud registration.
\newblock {\em IEEE Robotics and Automation Letters}, 2021.

\bibitem{parkison2018semantic}
Steven~A Parkison, Lu Gan, Maani~Ghaffari Jadidi, and Ryan~M Eustice.
\newblock Semantic iterative closest point through expectation-maximization.
\newblock In {\em BMVC}, page 280, 2018.

\bibitem{phan2018dgcnn}
Anh~Viet Phan, Minh Le~Nguyen, Yen Lam~Hoang Nguyen, and Lam~Thu Bui.
\newblock Dgcnn: A convolutional neural network over large-scale labeled
  graphs.
\newblock {\em Neural Networks}, 108:533--543, 2018.

\bibitem{pomerleau2015review}
Fran{\c{c}}ois Pomerleau, Francis Colas, and Roland Siegwart.
\newblock A review of point cloud registration algorithms for mobile robotics.
\newblock {\em Foundations and Trends in Robotics}, 4(1):1--104, 2015.

\bibitem{qi2017pointnet}
Charles~R Qi, Hao Su, Kaichun Mo, and Leonidas~J Guibas.
\newblock Pointnet: Deep learning on point sets for 3d classification and
  segmentation.
\newblock In {\em Proceedings of the IEEE conference on computer vision and
  pattern recognition}, pages 652--660, 2017.

\bibitem{qi2017pointnet++}
Charles~R Qi, Li Yi, Hao Su, and Leonidas~J Guibas.
\newblock Pointnet++: Deep hierarchical feature learning on point sets in a
  metric space.
\newblock {\em arXiv preprint arXiv:1706.02413}, 2017.

\bibitem{ramachandran2019stand}
Prajit Ramachandran, Niki Parmar, Ashish Vaswani, Irwan Bello, Anselm Levskaya,
  and Jonathon Shlens.
\newblock Stand-alone self-attention in vision models.
\newblock {\em arXiv preprint arXiv:1906.05909}, 2019.

\bibitem{rosen2020certifiably}
David~M Rosen, Luca Carlone, Afonso~S Bandeira, and John~J Leonard.
\newblock A certifiably correct algorithm for synchronization over the special
  euclidean group.
\newblock In {\em Algorithmic Foundations of Robotics XII}, pages 64--79.
  Springer, 2020.

\bibitem{rusinkiewicz2001efficient}
Szymon Rusinkiewicz and Marc Levoy.
\newblock Efficient variants of the icp algorithm.
\newblock In {\em Proceedings third international conference on 3-D digital
  imaging and modeling}, pages 145--152. IEEE, 2001.

\bibitem{rusu2009fast}
Radu~Bogdan Rusu, Nico Blodow, and Michael Beetz.
\newblock Fast point feature histograms (fpfh) for 3d registration.
\newblock In {\em 2009 IEEE international conference on robotics and
  automation}, pages 3212--3217. IEEE, 2009.

\bibitem{rusu2008aligning}
Radu~Bogdan Rusu, Nico Blodow, Zoltan~Csaba Marton, and Michael Beetz.
\newblock Aligning point cloud views using persistent feature histograms.
\newblock In {\em 2008 IEEE/RSJ international conference on intelligent robots
  and systems}, pages 3384--3391. IEEE, 2008.

\bibitem{salti2014shot}
Samuele Salti, Federico Tombari, and Luigi Di~Stefano.
\newblock Shot: Unique signatures of histograms for surface and texture
  description.
\newblock {\em Computer Vision and Image Understanding}, 125:251--264, 2014.

\bibitem{segal2009generalized}
Aleksandr Segal, Dirk Haehnel, and Sebastian Thrun.
\newblock Generalized-icp.
\newblock In {\em Robotics: science and systems}, volume~2, page 435. Seattle,
  WA, 2009.

\bibitem{serafin2015nicp}
Jacopo Serafin and Giorgio Grisetti.
\newblock Nicp: Dense normal based point cloud registration.
\newblock In {\em 2015 IEEE/RSJ International Conference on Intelligent Robots
  and Systems (IROS)}, pages 742--749. IEEE, 2015.

\bibitem{shi2021keypoint}
Chenghao Shi, Xieyuanli Chen, Kaihong Huang, Junhao Xiao, Huimin Lu, and Cyrill
  Stachniss.
\newblock Keypoint matching for point cloud registration using multiplex
  dynamic graph attention networks.
\newblock {\em IEEE Robotics and Automation Letters}, 2021.

\bibitem{tazir2018cicp}
M~Lamine Tazir, Tawsif Gokhool, Paul Checchin, Laurent Malaterre, and Laurent
  Trassoudaine.
\newblock Cicp: Cluster iterative closest point for sparse--dense point cloud
  registration.
\newblock {\em Robotics and Autonomous Systems}, 108:66--86, 2018.

\bibitem{vaswani2017attention}
Ashish Vaswani, Noam Shazeer, Niki Parmar, Jakob Uszkoreit, Llion Jones,
  Aidan~N Gomez, {\L}ukasz Kaiser, and Illia Polosukhin.
\newblock Attention is all you need.
\newblock In {\em Advances in neural information processing systems}, pages
  5998--6008, 2017.

\bibitem{wang2020axial}
Huiyu Wang, Yukun Zhu, Bradley Green, Hartwig Adam, Alan Yuille, and
  Liang-Chieh Chen.
\newblock Axial-deeplab: Stand-alone axial-attention for panoptic segmentation.
\newblock In {\em European Conference on Computer Vision}, pages 108--126.
  Springer, 2020.

\bibitem{wang2020linformer}
Sinong Wang, Belinda~Z Li, Madian Khabsa, Han Fang, and Hao Ma.
\newblock Linformer: Self-attention with linear complexity.
\newblock {\em arXiv preprint arXiv:2006.04768}, 2020.

\bibitem{wang2019deep}
Yue Wang and Justin~M Solomon.
\newblock Deep closest point: Learning representations for point cloud
  registration.
\newblock In {\em Proceedings of the IEEE/CVF International Conference on
  Computer Vision}, pages 3523--3532, 2019.

\bibitem{wang2019prnet}
Yue Wang and Justin~M Solomon.
\newblock Prnet: Self-supervised learning for partial-to-partial registration.
\newblock {\em arXiv preprint arXiv:1910.12240}, 2019.

\bibitem{wu20153d}
Zhirong Wu, Shuran Song, Aditya Khosla, Fisher Yu, Linguang Zhang, Xiaoou Tang,
  and Jianxiong Xiao.
\newblock 3d shapenets: A deep representation for volumetric shapes.
\newblock In {\em Proceedings of the IEEE conference on computer vision and
  pattern recognition}, pages 1912--1920, 2015.

\bibitem{yang2020teaser}
Heng Yang, Jingnan Shi, and Luca Carlone.
\newblock Teaser: Fast and certifiable point cloud registration.
\newblock {\em IEEE Transactions on Robotics}, 37(2):314--333, 2020.

\bibitem{yang2015go}
Jiaolong Yang, Hongdong Li, Dylan Campbell, and Yunde Jia.
\newblock Go-icp: A globally optimal solution to 3d icp point-set registration.
\newblock {\em IEEE transactions on pattern analysis and machine intelligence},
  38(11):2241--2254, 2015.

\bibitem{yew2020rpm}
Zi~Jian Yew and Gim~Hee Lee.
\newblock Rpm-net: Robust point matching using learned features.
\newblock In {\em Proceedings of the IEEE/CVF conference on computer vision and
  pattern recognition}, pages 11824--11833, 2020.

\bibitem{yuan2021tokens}
Li Yuan, Yunpeng Chen, Tao Wang, Weihao Yu, Yujun Shi, Zihang Jiang, Francis~EH
  Tay, Jiashi Feng, and Shuicheng Yan.
\newblock Tokens-to-token vit: Training vision transformers from scratch on
  imagenet.
\newblock {\em arXiv preprint arXiv:2101.11986}, 2021.

\bibitem{yuan2021volo}
Li Yuan, Qibin Hou, Zihang Jiang, Jiashi Feng, and Shuicheng Yan.
\newblock Volo: Vision outlooker for visual recognition, 2021.

\bibitem{yuan2020deepgmr}
Wentao Yuan, Benjamin Eckart, Kihwan Kim, Varun Jampani, Dieter Fox, and Jan
  Kautz.
\newblock Deepgmr: Learning latent gaussian mixture models for registration.
\newblock In {\em European Conference on Computer Vision}, pages 733--750.
  Springer, 2020.

\bibitem{zaheer2017deep}
Manzil Zaheer, Satwik Kottur, Siamak Ravanbakhsh, Barnabas Poczos, Ruslan
  Salakhutdinov, and Alexander Smola.
\newblock Deep sets.
\newblock {\em arXiv preprint arXiv:1703.06114}, 2017.

\bibitem{zhao2021point}
Hengshuang Zhao, Li Jiang, Jiaya Jia, Philip~HS Torr, and Vladlen Koltun.
\newblock Point transformer.
\newblock In {\em Proceedings of the IEEE/CVF International Conference on
  Computer Vision}, pages 16259--16268, 2021.

\bibitem{zhao2021centroidreg}
Hengwang Zhao, Zhidong Liang, Chunxiang Wang, and Ming Yang.
\newblock Centroidreg: A global-to-local framework for partial point cloud
  registration.
\newblock {\em IEEE Robotics and Automation Letters}, 6(2):2533--2540, 2021.

\bibitem{zhou2016fast}
Qian-Yi Zhou, Jaesik Park, and Vladlen Koltun.
\newblock Fast global registration.
\newblock In {\em European conference on computer vision}, pages 766--782.
  Springer, 2016.

\bibitem{zhou2018open3d}
Qian-Yi Zhou, Jaesik Park, and Vladlen Koltun.
\newblock Open3d: A modern library for 3d data processing.
\newblock {\em arXiv preprint arXiv:1801.09847}, 2018.

\end{thebibliography}
}

\end{document}